\documentclass{article}
\usepackage{xspace}
\newcommand{\modelname}{Cyclic NN\xspace}

\usepackage{ijcai23}

\usepackage{times}
\usepackage{soul}
\usepackage{url}
\usepackage[hidelinks]{hyperref}
\usepackage[utf8]{inputenc}
\usepackage[small]{caption}
\usepackage{graphicx}
\usepackage{amsmath}
\usepackage{amsthm}
\usepackage{booktabs}
\usepackage{amssymb}
\usepackage{algorithm}
\usepackage{algorithmic}
\usepackage[switch]{lineno}
\usepackage{multirow}
\usepackage{natbib}

\linenumbers

\title{Cyclic Neural Network}

\author{
Liangwei Yang$^1$
\and
Hengrui Zhang$^1$\and
Zihe Song$^1$\and
Jiawei Zhang$^2$\and
Jing Ma$^3$\and \\
Weizhi Zhang$^1$\And
Philip S. Yu$^1$
\affiliations
$^1$Department of Computer Science, University of Illinois Chicago, United States\\
$^2$IFM Lab, University of California, Davis, United States\\
$^3$University of Electronic Science and Technology of China, China\\
\emails
\{lyang84, hzhan55, zsong29, wzhan42, psyu\}@uic.edu,
jiwzhang@ucdavis.edu,
jingma@uestc.edu.cn
}

\begin{document}
\nolinenumbers

\maketitle

\begin{abstract}
This paper answers a fundamental question in artificial neural network (ANN) design: We do not need to build ANNs layer-by-layer sequentially to guarantee the Directed Acyclic Graph (DAG) property. Drawing inspiration from biological intelligence (BI), where neurons form a complex, graph-structured network, we introduce the groundbreaking Cyclic Neural Networks (Cyclic NNs). It emulates the flexible and dynamic graph nature of biological neural systems, allowing neuron connections in any graph-like structure, including cycles. This offers greater adaptability compared to the DAG structure of current ANNs. We further develop the Graph Over Multi-layer Perceptron, which is the first detailed model based on this new design paradigm. Experimental validation of the Cyclic NN's advantages on widely tested datasets in most generalized cases, demonstrating its superiority over current BP training methods through the use of a forward-forward (FF) training algorithm.
This research illustrates a totally new ANN design paradigm, which is a significant departure from current ANN designs, potentially leading to more biologically plausible AI systems.

\end{abstract}

\section{Introduction}
Artificial intelligence (AI) has reshaped our daily lives and is expected to have a much greater impact in the foreseeable future. Lying behind the most profound AI applications~\citep{silver2017mastering,gpt4,ramesh2021zero,jumper2021highly}, artificial neural networks (ANN) are designed specifically for different domains to fit the training data such as multi-layer perception (MLP)~\citep{rumelhart1986learning}, convolution neural network (CNN)~\citep{lecun1995convolutional} and Transformer~\citep{vaswani2017attention}. Regardless of the network structure, neural networks are stacked layer-by-layer to form deep ANNs for greater learning capacity. It has been a \textit{de facto
} practice until now that data is first fed into the input layer and then propagated through all the stacked layers to obtain the final representations at the output layer. In this paper, we seek to answer a fundamental question in ANNs: ``Do we really need to stack neural networks layer-by-layer sequentially?''.

\begin{figure}[!hbt]
    \centering
    \includegraphics[width=\linewidth]{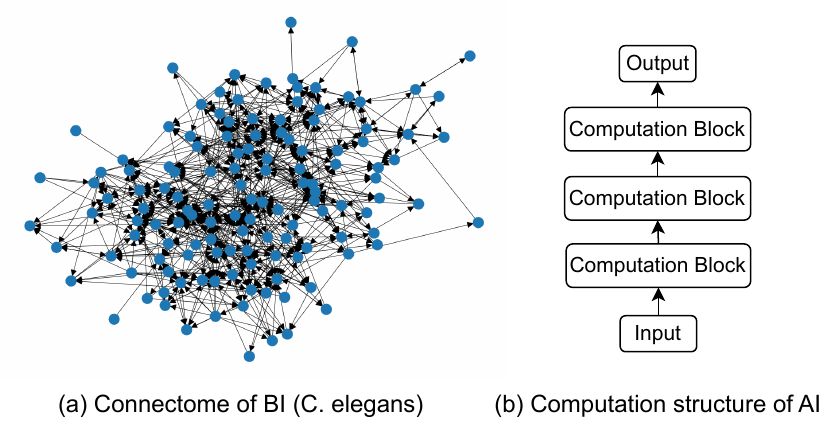}
    \caption{Neuron connection between Biology Neural Network and Artificial Neural Network}
    \label{BIvsAI}
\end{figure}

To answer this question, let's first examine the evidence from biological intelligence (BI). Neuroscientists have studied the biological neurons for decades. The connectome of C. elegans is the most thoroughly studied biological neural system, and biologists depicted the most detailed connection between 302 biological neurons~\citep{white1986structure,cook2019whole} as shown in Figure~\ref{BIvsAI}(a). Rather than being stacked layer-by-layer, all the neurons form a complicated connection graph, where each can connect to several other neurons within the system. We are even unable to determine which neuron serves as the input/output within the neural system to process information. The same findings have also been observed in the latter more complicated neural systems, such as the biology neural connectome of drosophila larva~\citep{winding2023connectome}, zebrafish~\citep{brooks2022mapping} and the mouse~\citep{sporns2014connections}. Observed biological intelligence exhibits graph-structured, flexible, and dynamic neural systems, which are apparently different from the current layer-by-layer AI models we build nowadays, as depicted in Figure~\ref{BIvsAI}(b).

To dive further, the difference in the neural system structure between BI and AI is actually caused by the learning rules. The Hebb's Rule~\citep{hebb2005organization}, depicted as ``Neurons that fire together wire together", is recognized as the fundamental learning way of biological neurons. The Spike-Timing-Dependent Plasticity (STDP) learning is then proposed to consider the relative spiking time of pre-synapse and post-synapse neurons further. Both learning rules of BI are localized, \textit{i.e.}, the learning occurs on each neuron within its local influence scope. The localized learning rules grant the flexibility of each neuron on its connections to other neurons, which leads to the complicated graph-structured BI system. Conversely, for AI systems, the backward propagation (BP) algorithm~\citep{rumelhart1986learning} has dominated the training of ANNs. Data is fed into the ANNs from the input layer, forward propagates layer by layer to the last layer, calculates a global loss for the whole ANN based on the ground-truth, and then reversely backward propagates the error signals layer by layer to the input layer. In this procedure, ANNs are trained by a global loss function and the ANNs must guarantee the error from global loss can be backpropagated layer by layer. This requirement prevent current ANNs from forming cycles for the ease of gradient back-propagation. As a result, most current ANNs are nearly all DAG structured.

To mitigate the biological implausible nature of the BP algorithm, the forward-forward (FF) algorithm~\citep{hinton2022forward} is recently proposed to train ANNs. FF algorithm constructs good/bad samples and computes a loss function on each layer to differentiate between these samples. Similar to Hebb's Rule and STDP learning, the FF algorithm is a localized learning method. 
This motivates us to conduct research on the Cyclic NN in this paper and to see whether the more complicated graph-structured ANNs can benefit AI systems. 
With cyclic structure within the neural network, Cyclic NN greatly increases the design space of ANNs beyond the DAG structure, and also largely enhances information communication between layers.


\modelname distinguishes itself from the current layer-by-layer ANNs in many aspects. 1) More flexible neuron connections. In \modelname, neurons can connect to each other to form any type of graph without any constraint. For example, the neurons can form a cycle connection as a loop. The flexible connection design makes \modelname more like the biological neural system. 2) Localized training. Instead of current dominating global loss-guided BP-based training, \modelname is based on localized training, \textit{i.e.}, each neuron is optimized with its own local loss function. There is no gradient propagating between neurons. 3) Computation. The neuron within \modelname is considered the computational neuron with greater computation capacity, such as a linear layer or a transformer block. It is inspired by the study of biological neuron~\citep{beniaguev2021single}, which empirically proves the learning capacity of a biological neuron is similar to that of an MLP. We take this observation and propose the computational neuron in \modelname with more complicated computation. In summary, our contributions can be summarized as follows:
\begin{itemize}
    \item Conceptually, we compare the difference between BI and AI to answer a fundamental question in ANN design: We do not need to organize ANNs in a layer-by-layer manner.
    \item Methodologically, we propose the ground-breaking Cyclic neural network, a novel ANN design paradigm that supports a much more flexible connection between neurons, which discards current layer-by-layer design constraints and is more biologically plausible.
    \item We test the novel design paradigm on the most generalized case and propose Graph Over Multi-layer Perceptron, the first detailed model based on Cyclic NN.
    \item Experimentally, we demonstrate the advantage of the proposed Cyclic neural network on widely tested datasets. At the same time, we are the first to beat the current dominating BP training using the FF training algorithm by the supported flexible network design proposed in this paper.
\end{itemize}

\section{Cyclic Neural Network}\label{NDANN}
In this section, we first illustrate the novel Cyclic neural network framework in detail and then discuss its tremendous advantages. One \modelname model is one graph $\mathcal{G}=(\mathcal{V}, \mathcal{E})$, where $\mathcal{V}=(N_1, N_2, ..., N_{|\mathcal{V}|})$ is the computational neuron set and $\mathcal{E}=(S_1, S_2, ..., S_{|\mathcal{E}|})$ is the synapse set denoting the connections among neurons. Similar to the BI system, $N_i$ ($\forall i \in \{1, 2, \cdots, |\mathcal{V}|\}$) is the neuron that tackles the detailed computation, and $S_j$ ($\forall j \in \{1, 2, \cdots, |\mathcal{E}|\}$) is the synapse that propagates information between computational neurons. In the \modelname, computational neurons can be connected flexibly in any way, like the BI system.

\subsection{Computational Neuron}~\label{neuron}
The computational neuron $N$ acts as the computation/optimization unit in Cyclic NN. Different from the current ANN in a neuron, it indicates a $d_{\text{input}}$ to $1$ mapping, we grant $N$ with stronger computation power. It is motivated by the research that proves the computation power of a single biological neuron is similar to an MLP~\citep{beniaguev2021single}. 
In \modelname, $N$ is parameterized by a function $f_{N}^{\mathbb{R}^{d_{\text{in}}^{N}} \rightarrow \mathbb{R}^{d_{\text{out}}^{N}}}(\mathbf{h}_{\text{in}}^{N}) = \mathbf{h}_{\text{out}}^{N}$ that maps from a $d_{\text{in}}^{N}$-dimensional representation $\mathbf{h}_{\text{in}}^{N}$ to a $d_{\text{out}}$-dimensional representation $\mathbf{h}_{\text{out}}^{N}$. 
$f_{N}$ can be parameterized by any computation block, such as MLP, CNN, LSTM, and Transformer. $d_{\text{in}}^{N}$ is the input dimension that is decided by the output of its pre-synapse neurons, and $d_{\text{out}}^{N}$ is the output dimension of $N$. Similar to the biological neurons, each computational neuron functions as a computation/optimization unit. Compared with current ANN neurons, the computational neuron is more independent during optimization.

\subsection{Synapse}
In neuroscience, synapses stand as pivotal junctions, orchestrating the complex symphony of neural communication. They serve as the critical interface between neurons, facilitating the transmission of information through chemical and electrical signals. In a \modelname, we model the synapse $S$ as the edge between neurons defined in Section~\ref{neuron}. Each synapse $S_{1,2} = (N_1 \to N_2)$ is a directional edge from computational neuron $N_1$ to $N_2$. It indicates the output of $N_1$, $\mathbf{h}_{\text{out}}^{N_1}$, will be propagated to $N_2$ as part of its input $\mathbf{h}_{\text{in}}^{N_2}$. Different from the current ANNs with DAG structure, the synapses between neurons can be organized as any connected graph structure, including the cyclic neural network.

\subsection{Local Optimization}
Local optimization is the bedrock to support \modelname, which is also a distinguishing point compared with current ANNs. For current ANNs, inputs are propagated through layers to obtain the final representation. Designing a global loss function $\mathcal{L}_{\text{global}}$ based on the final representation and ground-truth labels is the \textit{de facto} practice to train ANNs. $\mathcal{L}_{\text{global}}$ is optimized with BP algorithm to propagate the error signal layer-by-layer, which also prohibits the formation of computation cycle. Conversely, \modelname depends on local optimization, \textit{i.e.}, each computational neuron is optimized locally without gradients propagated from other computational neurons. For the computational neuron $N$, \modelname has a local loss function $\mathcal{L}_{\text{Local}}$ to optimize its parameters. The local optimization principle is similar to the BI system, where each neuron has the ability to learn from its local context.

\subsection{Inference}
During the inference phase, we design a readout layer to gather the output of all the neurons within the model as $\mathbf{h}_{\text{readout}}=f_{\text{readout}}([\mathbf{h}_{\text{out}}^{N_1}, \mathbf{h}_{\text{out}}^{N_2},..., \mathbf{h}_{\text{out}}^{N_{\mathcal{V}}}])$. $\mathbf{h}_{\text{readout}}$ collects all the encoded information and acts as the final representation for the inference task. For example, $\mathbf{h}_{\text{readout}}$ can be fed into a classifier for the classification task, where the classifier is trained together with the computational neurons.

\subsection{Advantages of \modelname}
\modelname has its unique advantages over current ANNs, which can be summarized as follows:
\begin{itemize}
    \item Flexibility: \modelname enables more flexible neuron connections. Neurons can be connected to form loops without any constraint.
    \item Extensibility: As computational neurons are more independent of other neurons, \modelname can be easily extended. We can insert a new computational neuron into the current trained \modelname without much impact.
    \item Parallelism: Due to the local optimization, each computational neuron can be optimized immediately when the data comes without the need to wait for the gradient to propagate back. It supports greater parallelism because all computational neurons can be optimized at the same time without waiting for a gradient.
    \item Privacy: In privacy-sensitive scenarios such as federated learning~\citep{he2021fedgraphnn}, the gradient leakage can lead to privacy issue. Without gradient propagated through layers, \modelname will not have privacy leakage issues.
    \item Biological Similarity: Compared with current ANNs, \modelname is more similar to biological intelligence. The design of computational neurons, synapses, and local optimization are all grounded in biological observation.
\end{itemize}

\begin{figure*}[!hbt]
    \centering
    \includegraphics[width=0.9\linewidth]{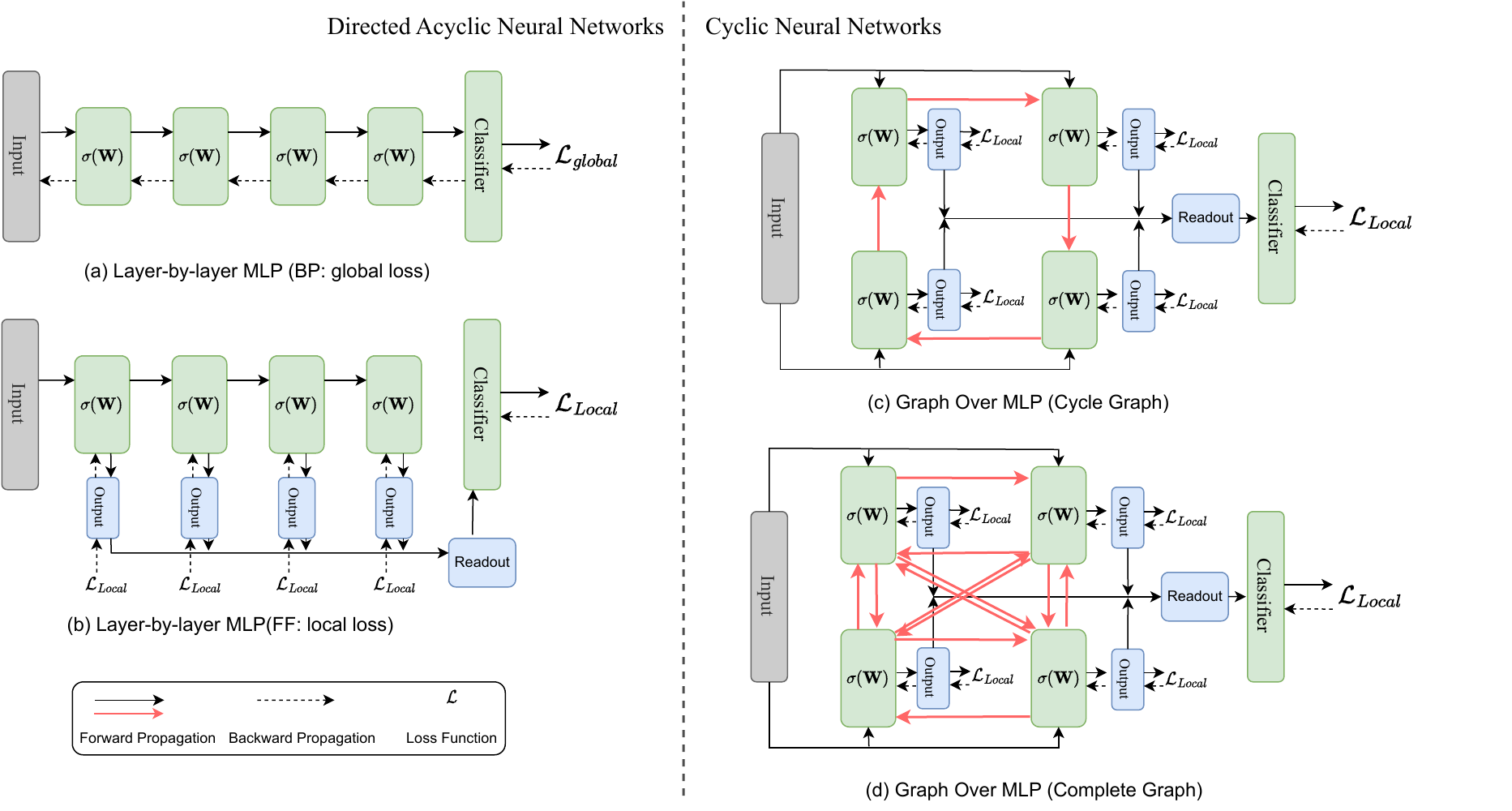}
    \caption{Comparison between different types of MLP structure.}
    \label{framework}
\end{figure*}

\section{Graph Over Multi-layer Perceptron}\label{GOMLP}

In this section, we propose the first \modelname under the most generalized case, Graph Over Multi-Layer Perceptron (GOMLP), to show the design principle of \modelname. As shown in Figure~\ref{framework}(c) and (d), GOMLP is designed by building a graph structure over the multi-layer perception to solve the classification task. GOMLP includes input construction, computation graph structure, synapse propagation, readout layer, optimization, and inference.

\subsection{Input Construction}\label{construction}
For the classification task, each sample is symbolized as the feature-label pair $(\mathbf{h}_i, y_i)$, where $\mathbf{h}_i$ is the representation of sample $i$ and $y_i$ is the corresponding label. To enable the local optimization illustrated in Section~\ref{optimization}, a fusion function is used to construct the input as:
\begin{align}\label{eq:input}
    \mathbf{h}_{\text{pos}} &= f_{\text{fusion}}(\mathbf{h}, \mathbf{y}_{\text{true}}) = \mathbf{h} || \mathbf{y}_{\text{true}}, \notag \\
    \mathbf{h}_{\text{neg}} &= f_{\text{fusion}}(\mathbf{h}, \mathbf{y}_{\text{false}}) = \mathbf{h}|| \mathbf{y}_{\text{false}}, \\
    \mathbf{h}_{\text{neu}} &= f_{\text{fusion}}(\mathbf{h}, \mathbf{y}_{\text{neutral}}) = \mathbf{h}|| \mathbf{y}_{\text{neu}} , \notag
\end{align}
$\mathbf{h}_{\text{pos}}$, $\mathbf{h}_{\text{neg}}$, and $\mathbf{h}_{\text{neu}}$ are the constructed input for local optimization of different parts. $f_{\text{fusion}}$ is a function to fuse information between feature and label, which is defined as a concat function ($||$) in this paper. $\mathbf{y}_{\text{true}}$ is the one-hot vector of ground-true label, $\mathbf{y}_{\text{false}}$ is the one-hot vector of a randomly sampled false label. For $\mathbf{y}_{\text{neu}}$, we place an $\frac{1}{\text{Class Number}}$ on all the dimensions of one-hot vector to indicate $\mathbf{h}_{\text{neutral}}$ is neutral to all classes. To be noted that the $f_{\text{fusion}}$ can be designed as any proper function to fuse information of the input feature and the label. In our study, we design it as a simple concat function same as~\citep{hinton2022forward}.


\subsection{Computation Graph}
The computation graph $\mathcal{G}$ contains the computational neurons $\mathcal{V}$ and the synapses $\mathcal{E}$. Each computational neuron $N \in \mathcal{V}$ is a local module for calculation and optimization, while each synapse $S$ defines how the information propagates between computational neurons. $\mathcal{G}$ can be defined as a graph generator:
\begin{equation}\label{eq:generator}
    \mathcal{G} = \text{Generator}(|\mathcal{V}|, |\mathcal{E}|).
\end{equation}
The above-generated graph $\mathcal{G}$ denotes a general graph structure. Meanwhile, to justify the effectiveness of the proposed Cyclic NN, we test multiple graph generators to build the graph.
\begin{itemize}
    \item Chain graph. Neurons are organized layer-by-layer as shown in Figure~\ref{framework}(b). In this case, GOMLP degrades to Hinton's method~\citep{hinton2022forward}.
    \item Cycle graph. Neurons form a cycle by connecting the neurons head-to-tail as shown in Figure~\ref{framework}(c).
    \item Complete graph. Each neuron connects to all the other neurons, as shown in Figure~\ref{framework}(d).
    \item Watts-Strogatz (WS) graph~\citep{watts1998collective}. It is a random graph generation model that produces graphs with small-world properties, including short average path lengths and high clustering.
    \item Barabási–Albert (BA) graph~\citep{albert2002statistical}. It generates random scale-free networks using a preferential attachment mechanism.
\end{itemize}

\subsubsection{Neuron Update}\label{sec:update}
In GOMLP, each neuron is parameterized by a linear layer. At each propagation, neuron $N$ is updated by (we omit the notation of $N$ in equation for simplicity): 
\begin{equation}\label{eq:update}
    \mathbf{h}_{\text{out}} = \sigma(\mathbf{W}\tilde{\mathbf{h}}_{\text{in}}),
\end{equation}
where $\sigma$ is the Relu activation function~\citep{nair2010rectified}, $\mathbf{W} \in \mathbb{R}_{d_{\text{out}}^{N} \times d_{\text{in}}^{N}}$ is $N$'s parameter. $d_{\text{out}}^{N}$ is $N$'s output dimension, which is a pre-defined dimension size, and $d_{\text{in}}^{N}$ is $N$'s input dimension, which is defined by the output of $N$'s pre-synapse neurons. $\tilde{\mathbf{h}}_{\text{in}}$ is the normalized input as:
\begin{equation}
    \tilde{\mathbf{h}}_{\text{in}} = \frac{\mathbf{h}_{\text{in}}}{\left\| \mathbf{h}_{\text{in}} \right\|_2},
\end{equation}
where $\mathbf{h}_{\text{in}}$ is computational neuron $N$'s input.

\subsubsection{Synapse Propagation}\label{sec:propagation}
Each synapse $S=(N_i \rightarrow N_j)$ is a directional edge from computational neuron $N_i$ to $N_j$, which indicates $N_i$ is the pre-synapse neuron of $N_j$ and $\mathbf{h}^{N_i}_{out}$ (the output of $N_i$) will be propagated to $N_j$. Assume for neuron $N$, we obtain a set of pre-synapse neurons $(N_1, N_2, ..., N_n)$ based on the topology of $\mathcal{G}$. Then, in each propagation, $N$ receives the output of all its pre-synapse neurons along the synapses, and fuse the information to form its input by a concatenation function:
\begin{equation}\label{eq:propagate}
    \mathbf{h}_{\text{in}} = \mathbf{h} || \mathbf{h}_{\text{out}}^{N_1} || \mathbf{h}_{\text{out}}^{N_2} ||, ..., || \mathbf{h}_{\text{out}}^{N_n},
\end{equation}
where $||$ is the concat function, $\mathbf{h}$ is the input representation constructed in Section~\ref{construction}. Then we can obtain $\mathbf{h}_{\text{in}, \text{pos}}$, $\mathbf{h}_{\text{in}, \text{neg}}$, $\mathbf{h}_{\text{in}, \text{neu}}$ by providing $\mathbf{h}_{\text{pos}}$, $\mathbf{h}_{\text{neg}}$, $\mathbf{h}_{\text{neu}}$ separately.
As we relax the layer-by-layer restriction, the differentiation between the input/hidden/output layers is also relaxed. In this case, we directly put the input $\mathbf{h}$ to all computational neurons. Thus, the input dimension size of $N$, $d_{\text{in}}^N = d_{\mathbf{h}} + d_{\text{out}}^{N_1}+d_{\text{out}}^{N_2}+...+d_{\text{out}}^{N_n}$. 

\subsection{Readout Layer}
The readout layer is designed to collect information from all computational neurons and make the collective decision on the studied classification task. The input of the readout layer is the concat function of all computational neurons:
\begin{equation}
    \mathbf{h}_{\text{in}}^{\text{readout}} = f_{\text{readout}}(\mathbf{h}_{\text{out}}^{N_*})= ||_{i=1}^{|\mathcal{V}|}(\mathbf{h}_{\text{out}}^{N_{i}}),
\end{equation}
where $||$ is the concat function. Then, the readout layer casts all the input representations to the output dimension:
\begin{equation}\label{eq:y_hat}
    \hat{\mathbf{y}} = \text{Softmax}(\mathbf{W}_{\text{readout}}\mathbf{h}_{\text{in}}^{\text{readout}}),
\end{equation}
where $\mathbf{W}_{\text{readout}} \in \mathbb{R}_{\text{Class Number} \times d(\mathbf{h}_{\text{in}}^{\text{readout}})}$ is the parameter of the readout layer and $\hat{\mathbf{y}}$ is the prediction vector on classes.

\subsection{Local Optimization}\label{optimization}
GOMLP comprises two parts that hold parameters: the computational neurons, and the readout layer. All parameters are optimized locally without gradient propagates between each parts. The computational neuron and readout layer are optimized differently with different inputs. The optimization is shown in Algorithm~\ref{alg:algorithm}, which contains the optimization of each computational neuron and the readout layer. The hyper-parameter $T$ indicates the number of synapse propagation over $\mathcal{G}$.

\subsubsection{Computational Neuron Optimization}
computational neurons are optimized to differentiate the positive examples from negative ones. For computational neuron $N$, its optimization involves $\mathbf{h}_{\text{in}, \text{pos}}$ and $\mathbf{h}_{\text{in}, \text{neg}}$. After the computational neuron update (Equation~\ref{eq:update}), we can get $\mathbf{h}_{\text{out}, \text{pos}}$ and $\mathbf{h}_{\text{out}, \text{neg}}$, respectively. Then, following~\citep{hinton2022forward}, a goodness function is used to calculate the goodness score on the representation:
\begin{equation}
    p(\mathbf{h}) = \sigma(\sum_i h_i^2 - \theta * d(\mathbf{h})),
\end{equation}
where $p(\mathbf{h})$ is the goodness score to judge $\mathbf{h}$, $d(\mathbf{h})$ is the dimension size of $\mathbf{h}$, $\sigma$ is the relu activation function and $\theta$ is the threshold hyper-parameter. The binary cross-entropy loss is used to optimize each computational neuron:
\begin{equation}~\label{eq:neuron_optimization}
    \mathcal{L}_{N}=-\frac{1}{|\mathcal{D}|}\sum_{\mathcal{D}}(\log(p(\mathbf{h_{\text{out},\text{pos}}}))-\log(p(\mathbf{h_{\text{out},\text{neg}}}))),
\end{equation}
where $\mathcal{D}$ is the dataset. The optimization of computational neurons aims to increase the neuron's output for positive samples while decreasing the neurons' output for negative samples. It enables each computational neuron its own ability to differentiate positive examples from negative ones.

\subsubsection{Readout Layer Optimization}
The readout layer is designed to accomplish the classification task for GOMLP. It reads the information from all computational neurons, and makes the decision over classes. To relieve the label leakage issue, the readout layer is only optimized with $\mathbf{h}_{\text{neutral}}$.
We use a multi-class cross-entropy loss to optimize the readout layer:
\begin{equation}~\label{eq:readout_optimization}
    \mathcal{L}_{\text{Readout}}(\mathbf{y}, \hat{\mathbf{y}})= -\frac{1}{|\mathcal{D}|}\sum_{|\mathcal{D}|}\sum_{c=1}^{C} y_{c} \log(\hat{y_{c}}),
\end{equation}
where $C$ is the number of classes, $\mathbf{y}$ is the one-hot vector of ground-truth label and $\hat{\mathbf{y}}$ is the prediction from Equation~\ref{eq:y_hat}.
Though the optimization of the computational neuron and readout layer are localized and different, these two parts complement each other. Computational neurons aim to extract the hidden representations of each sample, and the readout layer aims to make the final decision based on the extracted information. 

During the inference time, we pair each test sample with the neutral label to construct $\mathbf{h}_{\text{neu}}$. It then propagates through the GOMLP to obtain its representation on each computational neuron. Finally, we predict its class with the largest logit from the output of the readout layer.


\begin{algorithm}[tb]
    \caption{Graph Over MLP}
    \label{alg:algorithm}
    \textbf{Input}:  dataset $\mathcal{D}$\\
    \textbf{Parameter}: $\mathcal{G} = (\mathcal{V}, \mathcal{E})$, $\mathbf{W}_{\text{readout}}$ \\
    \textbf{Output}: Optimized $\mathcal{G}$, $\mathbf{W}_{\text{readout}}$
    \begin{algorithmic}[1] 
        \WHILE{\textit{Not Converged}}
        \STATE Obtain inputs by Equation~\ref{eq:input} from $\mathcal{D}$ \\
        \STATE Let $t=0$ \\
        \WHILE{$t < T$}
        \FOR{$N \in \mathcal{V}$}
        \STATE Synapse Propagate by Eq.~\ref{eq:propagate}
        \STATE computational neuron Update by Eq.~\ref{eq:update}
        \STATE Optimize $N$ by Eq.~\ref{eq:neuron_optimization}
        \ENDFOR
        \STATE $t = t + 1$ \\
        \ENDWHILE
        \STATE Calculate the output of Readout layer by Eq.~\ref{eq:p}
        \STATE Optimize $\mathbf{W}_{\text{readout}}$ by Eq.~\ref{eq:readout_optimization}
        \ENDWHILE
        \RETURN $\mathcal{G} = (\mathcal{V}, \mathcal{E})$, $\mathbf{W}_{\text{readout}}$
    \end{algorithmic}
\end{algorithm}

\section{Experiments}\label{Experiments}
In this section, we conduct experiments on three datasets to test the performance and characteristics of GOMLP.

\subsection{Experimental setup}
\subsubsection{Datasets}
We conduct experiments on three widely studied datasets from both computer vision and natural language processing domains. Data statistics are shown in Table~\ref{tab:dataset}. For each dataset, the split of training and test follows the original setting. We further extract $20\%$ samples from the training data as validation sets to tune hyper-parameters.
\begin{itemize}
    \item MNIST~\citep{lecun1989handwritten}. It contains handwritten digits from 0-9, which is the most accessible and used datasets in the field of machine learning.
    \item NewsGroup~\citep{lang1995newsweeder}. It is a collection of approximately 20,000 newsgroup documents, partitioned across 20 different newsgroups. This dataset is widely used for experiments in text applications of machine learning techniques, such as text classification and text clustering.
    \item IMDB~\citep{maas2011learning}. It is a movie review dataset crawled from IMDB. It is the most widely studied dataset for binary sentiment classification.
\end{itemize}
For MNIST, we directly use its flattened pixel values as the input of all methods and replace the first $10$ pixels with labels as the fusion function, which is the same as~\citep{hinton2022forward} and leads to an input dimension of $28 * 28 = 784$. For NLP datasets (NewsGroup, IMDB), we use BERT~\citep{devlin2018bert} to encode the sentences into a fixed-length tensor ($768$) as the input. The fusion function is the concat function, which leads to an input dimension of $768+20=788$ for NewsGroup and $768+2=770$ for IMDB dataset, respectively.

\begin{table}[htb]
\caption{Dataset Statistics}
\label{tab:dataset}
\begin{tabular}{llccc}
\toprule
Dataset & MNIST & NewsGroup & IMDB  \\
\midrule
Training Samples & 50,000 & 9,314 & 20,000 \\
Validation Samples & 10,000 & 2,000 & 5,000 \\
Test Samples & 10,000 & 7,532 & 25,000 \\
Dimensions & 784 & 788 & 770 \\
Classes & 10 & 20 & 2 \\
\hline
\bottomrule
\end{tabular}
\end{table}

\subsubsection{Baselines}
In this paper, we aim to reveal the advantages of graph-structured multi-layer perceptron.
We compared GOMLP with a variant of different methods, which can be differentiated by two attributes (Training and Graph). Training indicates the training method, where BP indicates Backward Propagation~\citep{rumelhart1986learning} and FF indicates the Forward-forward algorithm~\citep{hinton2022forward}. The graph indicates the graph structure of computational neurons. To make a fair comparison, we keep $4$ layers of MLP for all methods. The special cases are further illustrated as:
\begin{itemize}
    \item BP-Chain*: Layer-by-layer neural networks trained with BP as depicted in Figure~\ref{framework}(a). It is the current default way of building and training ANNs.
    \item FF-Chain: Layer-by-layer neural networks trained with FF as depicted in Figure~\ref{framework}(b). This setting is the same as~\citep{hinton2022forward}.
    \item BP-Chain: A modified version of BP-Chain*, where we use the structure of Figure~\ref{framework}(b) with readout function and trained with BP. It adds direct local supervision on each layer.
\end{itemize}
FF-Cycle, FF-WSGraph, FF-BAGraph, and FF-Complete are different versions of GOMLP, where the training is FF and only the graph generator defined in Eq.~\ref{eq:generator} differs.

\subsubsection{Experimental Setting}
We use Adam~\citep{kingma2014adam} optimizer to train the model until convergence for all experiments. The learning rate and weight decay are tuned within (0.1,0.01,0.001) and (0.0, 1e-2, 1e-4, 1e-6, 1e-8), respectively. The early stop technique is applied to avoid overfitting, where we stop training if there is no improvement on the validation set for continuous $10$ epochs. For each setting, we report the mean and variance on $20$ experiments with different random seed.

\subsection{Overall Comparison}

\begin{table}[htb]
\caption{Error rate (\%) $\downarrow$ on different datasets.}
\label{Comparison Experiment}
\begin{tabular}{llccc}
\toprule
Train & Graph & MNIST & NewsGroup & IMDB \\
\midrule
BP & Chain* & $1.77_{\pm 0.16}$ & $42.11_{\pm0.92}$ & $\textbf{17.16}_{\pm 0.19}$  \\
FF & Chain & $1.83_{\pm 0.2}$ & $43.88_{\pm 0.38}$ & $18.75_{\pm 0.92}$  \\
BP & Chain & $1.74_{\pm 0.11}$ & $38.85_{\pm 0.42}$ & $17.27_{\pm 0.13}$  \\
\hline 
FF & Cycle & $1.80_{\pm 0.14}$ & $43.54_{\pm 0.41}$ & $18.97_{\pm 0.49}$ \\
FF & WSGraph & $1.70_{\pm 0.17}$ & $38.38_{\pm 0.13}$ & $17.93_{\pm 0.38}$  \\
FF & BAGraph & $1.64_{\pm 0.08}$ & $38.41_{\pm 0.14}$ & $18.20_{\pm 0.67}$  \\
FF & Complete& $\textbf{1.54}_{\pm 0.05}$ & $\textbf{38.366}_{\pm 0.06}$ & $17.58_{\pm 0.20}$  \\
\hline
\bottomrule
\end{tabular}
\end{table}

The overall experiment result is shown in Table~\ref{Comparison Experiment}. We show the error rate of different methods on different datasets (the lower, the better). Best performance is marked bold. From the table, we can have several interesting and exciting findings:
\begin{itemize}
    \item FF-Complete achieves the best performance on MNIST and NewsGroup datasets, and it also achieves comparable results to the best one on the IMDB dataset. It is the first FF-trained model that outcompetes the BP-trained model. It is an exciting observation of the effectiveness of the FF algorithm compared with the BP algorithm. 
    \item FF-Chain performs worse than BP-Chain* on all datasets. This observation is on par with~\citep{hinton2022forward}, where the FF lags behind the BP training algorithm when they both follow layer-by-layer organization as a chain graph. However, we can outcompete BP-Chain* when we organize the layers as a graph structure. This finding inevitably reveals the advantages of GOMLP by organizing multi-layer perceptron as a flexible graph structure.
    \item FF-Cycle achieves similar performance with FF-Chain on three datasets. It is reasonable because there is only one edge difference between these two methods. When we build more complex graphs (WSGraph, BAGraph, Complete Graph), we can observe much better performance immediately. It shows the benefits of enriching the communication between layers by the GOMLP.
    \item BP-Chain is better than BP-Chain* in most cases. Compared with BP-Chain*, BP-Chain further adds layer-wise optimization directly from the final loss function. It indicates the advantageous layer-wise optimization, which provides new guidelines when designing layer-by-layer neural networks.
\end{itemize}
In summary, the experiment results answer that we do not need to stack neural networks layer-by-layer sequentially, and we can organize the neural networks as a flexible, complex graph structure like the brain. More excitingly, we can outperform the current \textit{de facto} layer-by-layer neural network design paradigm with the cyclic neural network, and provide a totally new way of building ANNs.

\subsection{Hyper-paramter Sensitiviey}

\begin{figure}[t]
    \begin{center}
    \includegraphics[width=.235\textwidth]{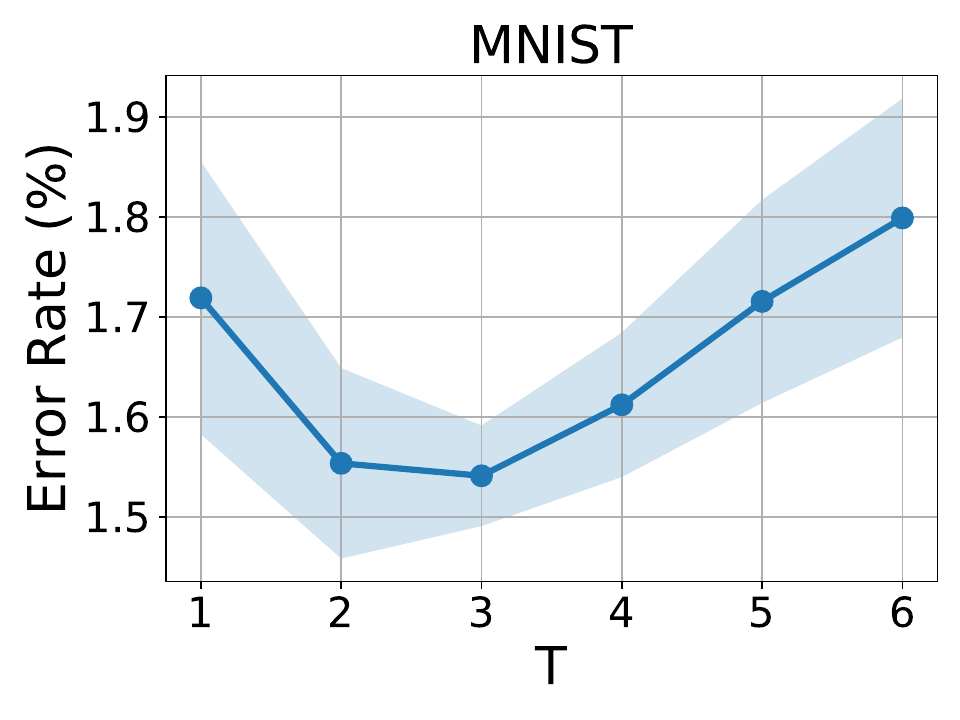}
    \includegraphics[width=.235\textwidth]{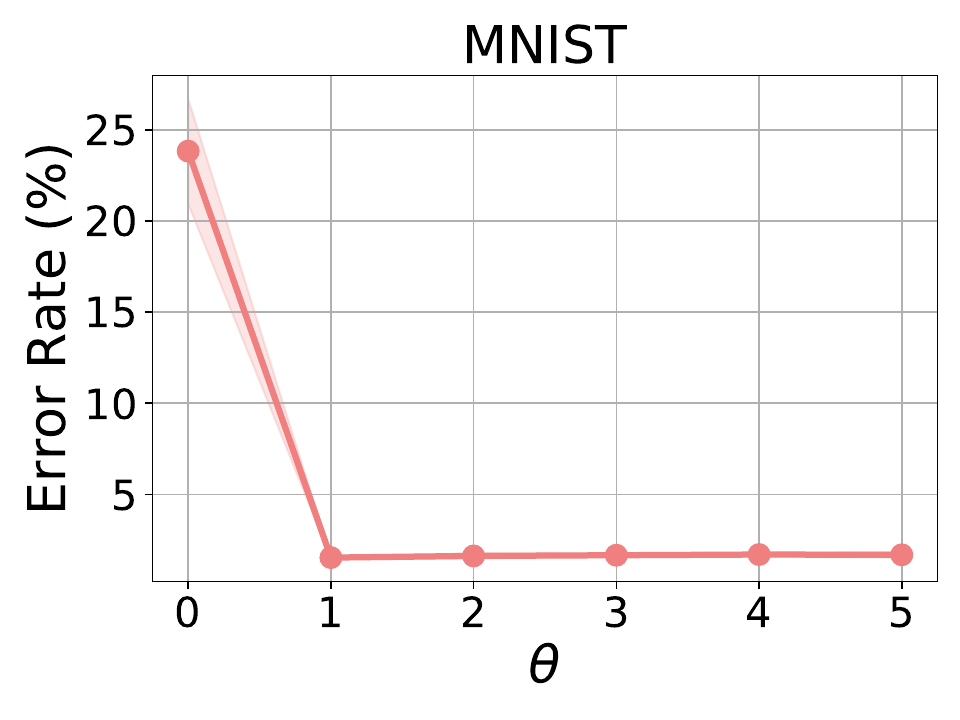}
    \includegraphics[width=.235\textwidth]{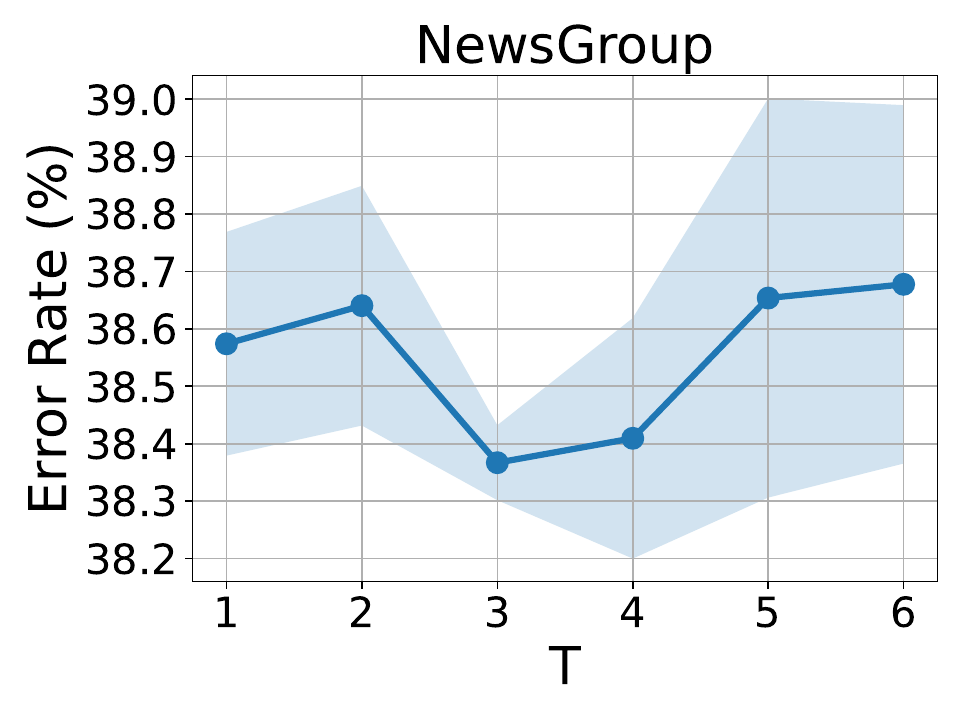}
    \includegraphics[width=.235\textwidth]{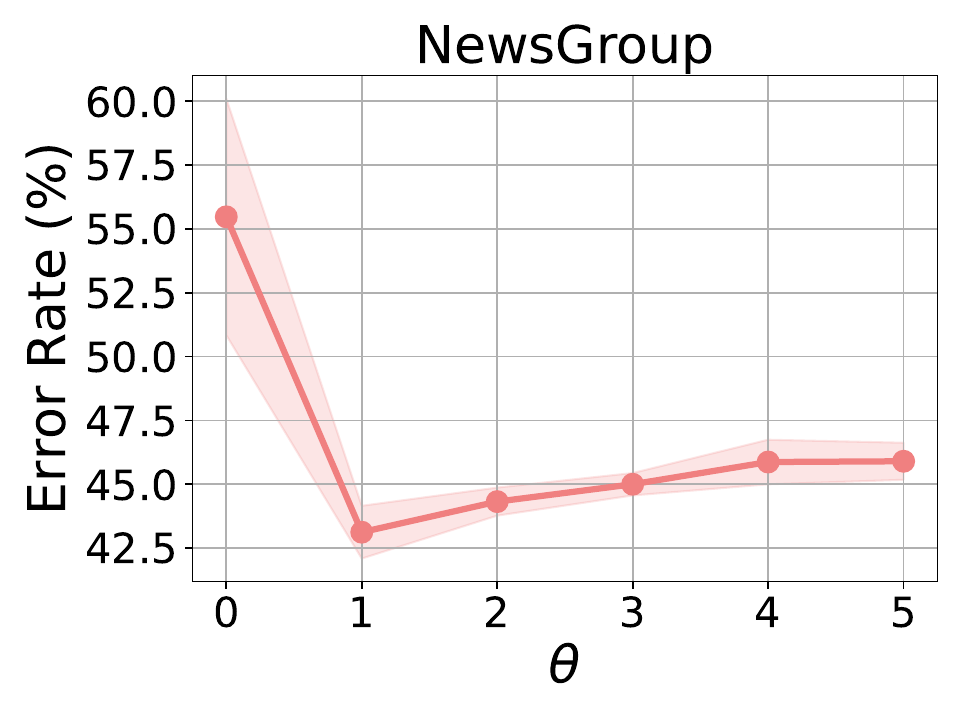}
    \includegraphics[width=.235\textwidth]{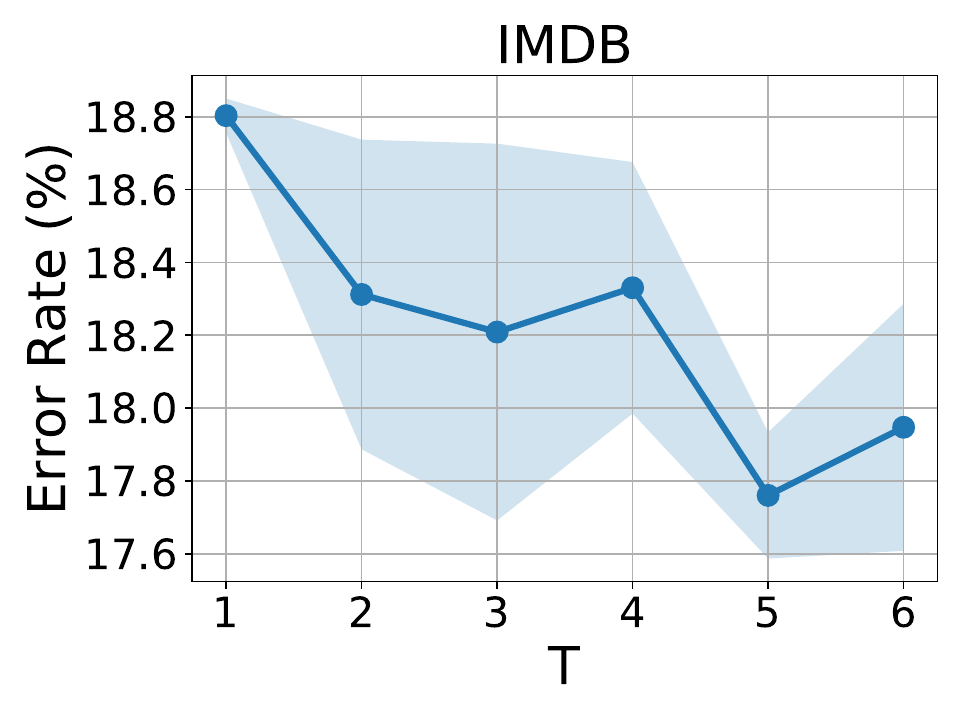}
    \includegraphics[width=.235\textwidth]{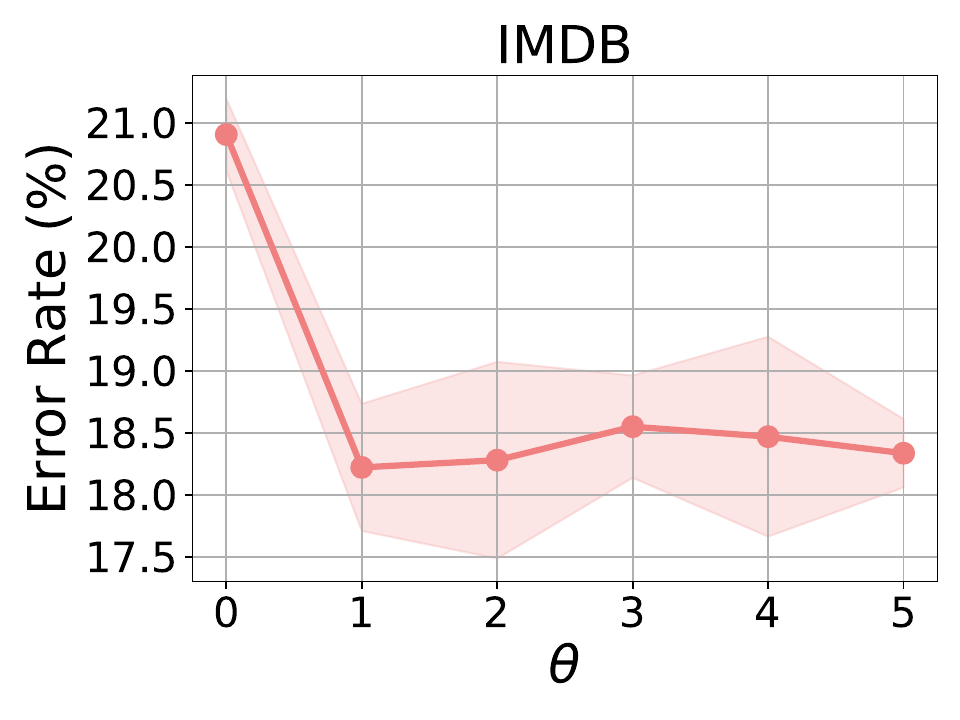}
    \end{center}
    \caption{Parameter sensitivity of $T$ and $\theta$}
    \label{fig:hyper-parameter}
\end{figure}

In this section, we test the impact of hyper-parameters ($T$ and $\theta$) designed within GOMLP. Experiment results on three datasets are shown in Figure~\ref{fig:hyper-parameter}. 

$T$ controls the number of propagation between computational neurons. A larger $T$ indicates more times the information is propagated between computational neurons. We can observe an error rate trend that first decreases and then increases on all three datasets. It indicates that the computational neurons need a suitable propagation number. When $T$ is small, computational neurons can not draw sufficient lessons from each other. When $T$ is large, computational neurons are over-propagated, which leads to the over-smoothing problem. Different from over-smoothing in graph neural network~\citep{chen2020measuring} where the graph is data and the over-smoothed node representation, in GOMLP, the graph is the model, and the over-smoothing occurs among computational neurons.

$\theta$ controls the goodness threshold of each computational neuron. We can observe a sharp error rate decrease when $\theta$ increases from $0$ to $1$, and then it gets stable with larger $\theta$. It indicates the existence of the goodness threshold matters more than the threshold value. When $\theta=0$, there is little room to optimize the computational neuron towards the negative sample, which can lead to the training collapse as the computational neuron can not differentiate the negative sample. When $\theta$ is larger, there is more room to optimize the goodness score toward a negative sample, as all the goodness scores under the threshold can represent a negative sample. $\theta$ is a critical component of GOMLP.

\subsection{Ablation Study}

\begin{table}[htb]
\caption{Error rate (\%) $\downarrow$ of Ablation study.}
\label{tab:ablation}
\begin{tabular}{lccc}
\toprule
Model & MNIST & NewsGroup & IMDB \\
\midrule
FF-Complete & \textbf{1.54} & \textbf{38.36} & \textbf{18.20} \\
$\quad$-$\mathcal{L}_{N}$ & 2.34 & 47.61 & 22.94 \\
$\quad$-$\mathcal{L}_{\text{Readout}}$ & 95.58 & 95.55 & 44.36 \\
\hline
\bottomrule
\end{tabular}
\end{table}

This section studies the impact of different optimization modules within GOMLP, including the computational neuron optimization $\mathcal{L}_{N}$ and readout layer optimization $\mathcal{L}_{\text{Readout}}$. We conduct experiments on the FF-Complete structure, and the results are summarized in Table~\ref{tab:ablation}. We can have the following observations: 1) The error rate increases when removing any optimization module, indicating the usefulness of each component. 2) GOMLP falls to a very large error rate (nearly random guess) when removing $\mathcal{L}_{\text{Readout}}$. It is reasonable as we depend on the readout layer to complete the final classification task. Without optimization on the readout layer, GOMLP falls into random guess even with optimized computational neuron's input. 3) The error rate increases by removing $\mathcal{L}_{N}$. It shows the computational neuron's optimization can provide a more informative goodness score for the readout layer to complete the classification task. $\mathcal{L}_{N}$ and $\mathcal{L}_{\text{Readout}}$ complement each other within GOMLP, and they collectively make the best performance.


\section{Related Work}\label{Related work}

\subsection{Localized Learning Algorithm}
Although end-to-end backpropagation has become the dominant training algorithm for deep neural networks, studies have revealed notable limitations in such end-to-end training with global objectives \citep{bengio2015towards,Crick1989TheRE}. Numerous works have proposed alternative training methods to make ANNs more biologically plausible. Inspired by Hebbian theory \citep{hebb2005organization}, Hebbian Learning \citep{Gerstner2002MathematicalFO} updates weights locally between two active, connected neurons, ensuring long-term stability so previously learned information is not lost. Addressing the weight transportation problem of backpropagation when applied to biological neurons, Feedback alignment methods \citep{lillicrap2016random,nokland2016direct} replace downstream synaptic weights with random weights, eliminating the need for feedback weights in neurons. Unlike backpropagation, which requires two types of computations in forward and backward passes, Equilibrium Propagation \citep{scellier2017equilibrium} performs both inference and weight updates using only one type of computation. The approach in \citep{pmlr-v97-nokland19a} reduces memory consumption and increases training parallelism by adopting subnetworks and layer-wise training. Additionally, \citep{hinton2022forward} introduces a simple yet efficient local objective function that measures the goodness of positive and negative data, along with a local optimization method to train ANNs.

Localized learning algorithm is the bedrock to support the cyclic structure within neural network. In this paper, we are the first to beat global BP training with pure localized learning algorithm by the help of cyclic network structure proposed in this paper.

\subsection{Graph Generator}
With a number of computational neurons as nodes, there are plenty of algorithms to generate computation graphs. The simplest model to generate a random graph is the Erdős–Rényi model \citep{erdds1959random}, which iterates all possible edges and adds them with a probability of $p \in [0,1]$. \citep{Batagelj2005EfficientGO} propose a more efficient algorithm to create a graph with a linear computation cost in terms of running time and space requirement. Random Geometric graphs \citep{penrose2003random} place nodes on geometric planes and form edges within a fixed distance, resembling real-world social networks. The Spectrum Graph Forge \citep{10.1109/INFOCOM.2018.8485916} creates a random graph that has the same eigenstructure as a given adjacency matrix. The Waxman model \citep{12889} generates large-scale graphs that have properties similar to those of communication networks. K-regular graphs \citep{STEGER_WORMALD_1999} are simple graphs in which every node has the same degree. The Watts-Strogatz model \citep{watts1998collective} produces small-world graphs that guarantee high clustering coefficients and low averages of shortest path distances. The Barabási-Albert model \citep{albert2002statistical} generates graphs by continuously adding nodes that attach to existing nodes with high degrees, known as scale-free graphs.

Due to the constraint of BP training algorithm, current ANNs only fall into a very specialized case considering the graph structure (Directed Acyclic Graph). In this paper, we propose a ground-breaking cyclic NN design paradigm to support all kinds of graph structure, which is more flexible and with higher biology similarity.

\section{Conclusion}
In summary, this research introduces Cyclic Neural Networks (Cyclic NNs), a novel ANN architecture inspired by the complex, graph-like neural networks in biological intelligence. This groundbreaking design diverges from traditional directed acyclic ANN structures. Our findings, demonstrated through the Graph Over Multi-layer Perceptron model and validated on various datasets, show enhanced performance over conventional backpropagation methods. This significant development paves the way for more efficient and biologically realistic AI systems, representing a major shift in ANN design.






\bibliographystyle{named}
\bibliography{ijcai23}

\begin{thebibliography}{}

\bibitem[\protect\citeauthoryear{Albert and Barab{\'a}si}{2002}]{albert2002statistical}
R{\'e}ka Albert and Albert-L{\'a}szl{\'o} Barab{\'a}si.
\newblock Statistical mechanics of complex networks.
\newblock {\em Reviews of modern physics}, 74(1):47, 2002.

\bibitem[\protect\citeauthoryear{Baldesi \bgroup \em et al.\egroup }{2018}]{10.1109/INFOCOM.2018.8485916}
Luca Baldesi, Carter~T. Butts, and Athina Markopoulou.
\newblock Spectral graph forge: Graph generation targeting modularity.
\newblock In {\em IEEE INFOCOM 2018 - IEEE Conference on Computer Communications}, page 1727–1735. IEEE Press, 2018.

\bibitem[\protect\citeauthoryear{Batagelj and Brandes}{2005}]{Batagelj2005EfficientGO}
Vladimir Batagelj and Ulrik Brandes.
\newblock Efficient generation of large random networks.
\newblock {\em Physical review. E, Statistical, nonlinear, and soft matter physics}, 71 3 Pt 2A:036113, 2005.

\bibitem[\protect\citeauthoryear{Bengio \bgroup \em et al.\egroup }{2015}]{bengio2015towards}
Yoshua Bengio, Dong-Hyun Lee, Jorg Bornschein, Thomas Mesnard, and Zhouhan Lin.
\newblock Towards biologically plausible deep learning.
\newblock {\em arXiv preprint arXiv:1502.04156}, 2015.

\bibitem[\protect\citeauthoryear{Beniaguev \bgroup \em et al.\egroup }{2021}]{beniaguev2021single}
David Beniaguev, Idan Segev, and Michael London.
\newblock Single cortical neurons as deep artificial neural networks.
\newblock {\em Neuron}, 109(17):2727--2739, 2021.

\bibitem[\protect\citeauthoryear{Brooks \bgroup \em et al.\egroup }{2022}]{brooks2022mapping}
Paul Brooks, Andrew Champion, and Marta Costa.
\newblock Mapping of the zebrafish brain takes shape.
\newblock {\em Nature Methods}, 19(11):1345--1346, 2022.

\bibitem[\protect\citeauthoryear{Chen \bgroup \em et al.\egroup }{2020}]{chen2020measuring}
Deli Chen, Yankai Lin, Wei Li, Peng Li, Jie Zhou, and Xu~Sun.
\newblock Measuring and relieving the over-smoothing problem for graph neural networks from the topological view.
\newblock In {\em Proceedings of the AAAI conference on artificial intelligence}, volume~34, pages 3438--3445, 2020.

\bibitem[\protect\citeauthoryear{Cook \bgroup \em et al.\egroup }{2019}]{cook2019whole}
Steven~J Cook, Travis~A Jarrell, Christopher~A Brittin, Yi~Wang, Adam~E Bloniarz, Maksim~A Yakovlev, Ken~CQ Nguyen, Leo T-H Tang, Emily~A Bayer, Janet~S Duerr, et~al.
\newblock Whole-animal connectomes of both caenorhabditis elegans sexes.
\newblock {\em Nature}, 571(7763):63--71, 2019.

\bibitem[\protect\citeauthoryear{Crick}{1989}]{Crick1989TheRE}
Francis Crick.
\newblock The recent excitement about neural networks.
\newblock {\em Nature}, 337:129--132, 1989.

\bibitem[\protect\citeauthoryear{Devlin \bgroup \em et al.\egroup }{2018}]{devlin2018bert}
Jacob Devlin, Ming-Wei Chang, Kenton Lee, and Kristina Toutanova.
\newblock Bert: Pre-training of deep bidirectional transformers for language understanding.
\newblock {\em arXiv preprint arXiv:1810.04805}, 2018.

\bibitem[\protect\citeauthoryear{ERDdS and R\&wi}{1959}]{erdds1959random}
P~ERDdS and A~R\&wi.
\newblock On random graphs i.
\newblock {\em Publ. math. debrecen}, 6(290-297):18, 1959.

\bibitem[\protect\citeauthoryear{Gerstner and Kistler}{2002}]{Gerstner2002MathematicalFO}
Wulfram Gerstner and Werner~M. Kistler.
\newblock Mathematical formulations of hebbian learning.
\newblock {\em Biological Cybernetics}, 87:404--415, 2002.

\bibitem[\protect\citeauthoryear{He \bgroup \em et al.\egroup }{2021}]{he2021fedgraphnn}
Chaoyang He, Keshav Balasubramanian, Emir Ceyani, Carl Yang, Han Xie, Lichao Sun, Lifang He, Liangwei Yang, Philip~S Yu, Yu~Rong, et~al.
\newblock Fedgraphnn: A federated learning system and benchmark for graph neural networks.
\newblock {\em arXiv preprint arXiv:2104.07145}, 2021.

\bibitem[\protect\citeauthoryear{Hebb}{2005}]{hebb2005organization}
Donald~Olding Hebb.
\newblock {\em The organization of behavior: A neuropsychological theory}.
\newblock Psychology press, 2005.

\bibitem[\protect\citeauthoryear{Hinton}{2022}]{hinton2022forward}
Geoffrey Hinton.
\newblock The forward-forward algorithm: Some preliminary investigations.
\newblock {\em arXiv preprint arXiv:2212.13345}, 2022.

\bibitem[\protect\citeauthoryear{Jumper \bgroup \em et al.\egroup }{2021}]{jumper2021highly}
John Jumper, Richard Evans, Alexander Pritzel, Tim Green, Michael Figurnov, Olaf Ronneberger, Kathryn Tunyasuvunakool, Russ Bates, Augustin {\v{Z}}{\'\i}dek, Anna Potapenko, et~al.
\newblock Highly accurate protein structure prediction with alphafold.
\newblock {\em Nature}, 596(7873):583--589, 2021.

\bibitem[\protect\citeauthoryear{Kingma and Ba}{2014}]{kingma2014adam}
Diederik~P Kingma and Jimmy Ba.
\newblock Adam: A method for stochastic optimization.
\newblock {\em arXiv preprint arXiv:1412.6980}, 2014.

\bibitem[\protect\citeauthoryear{Lang}{1995}]{lang1995newsweeder}
Ken Lang.
\newblock Newsweeder: Learning to filter netnews.
\newblock In {\em Machine learning proceedings 1995}, pages 331--339. Elsevier, 1995.

\bibitem[\protect\citeauthoryear{LeCun \bgroup \em et al.\egroup }{1989}]{lecun1989handwritten}
Yann LeCun, Bernhard Boser, John Denker, Donnie Henderson, Richard Howard, Wayne Hubbard, and Lawrence Jackel.
\newblock Handwritten digit recognition with a back-propagation network.
\newblock {\em Advances in neural information processing systems}, 2, 1989.

\bibitem[\protect\citeauthoryear{LeCun \bgroup \em et al.\egroup }{1995}]{lecun1995convolutional}
Yann LeCun, Yoshua Bengio, et~al.
\newblock Convolutional networks for images, speech, and time series.
\newblock {\em The handbook of brain theory and neural networks}, 3361(10):1995, 1995.

\bibitem[\protect\citeauthoryear{Lillicrap \bgroup \em et al.\egroup }{2016}]{lillicrap2016random}
Timothy~P Lillicrap, Daniel Cownden, Douglas~B Tweed, and Colin~J Akerman.
\newblock Random synaptic feedback weights support error backpropagation for deep learning.
\newblock {\em Nature communications}, 7(1):13276, 2016.

\bibitem[\protect\citeauthoryear{Maas \bgroup \em et al.\egroup }{2011}]{maas2011learning}
Andrew Maas, Raymond~E Daly, Peter~T Pham, Dan Huang, Andrew~Y Ng, and Christopher Potts.
\newblock Learning word vectors for sentiment analysis.
\newblock In {\em Proceedings of the 49th annual meeting of the association for computational linguistics: Human language technologies}, pages 142--150, 2011.

\bibitem[\protect\citeauthoryear{Nair and Hinton}{2010}]{nair2010rectified}
Vinod Nair and Geoffrey~E Hinton.
\newblock Rectified linear units improve restricted boltzmann machines.
\newblock In {\em Proceedings of the 27th international conference on machine learning (ICML-10)}, pages 807--814, 2010.

\bibitem[\protect\citeauthoryear{N{\o}kland and Eidnes}{2019}]{pmlr-v97-nokland19a}
Arild N{\o}kland and Lars~Hiller Eidnes.
\newblock Training neural networks with local error signals.
\newblock In Kamalika Chaudhuri and Ruslan Salakhutdinov, editors, {\em Proceedings of the 36th International Conference on Machine Learning}, volume~97 of {\em Proceedings of Machine Learning Research}, pages 4839--4850. PMLR, 09--15 Jun 2019.

\bibitem[\protect\citeauthoryear{N{\o}kland}{2016}]{nokland2016direct}
Arild N{\o}kland.
\newblock Direct feedback alignment provides learning in deep neural networks.
\newblock {\em Advances in neural information processing systems}, 29, 2016.

\bibitem[\protect\citeauthoryear{OpenAI}{2023}]{gpt4}
OpenAI.
\newblock {GPT-4} technical report.
\newblock {\em CoRR}, abs/2303.08774, 2023.

\bibitem[\protect\citeauthoryear{Penrose}{2003}]{penrose2003random}
Mathew Penrose.
\newblock {\em Random geometric graphs}, volume~5.
\newblock OUP Oxford, 2003.

\bibitem[\protect\citeauthoryear{Ramesh \bgroup \em et al.\egroup }{2021}]{ramesh2021zero}
Aditya Ramesh, Mikhail Pavlov, Gabriel Goh, Scott Gray, Chelsea Voss, Alec Radford, Mark Chen, and Ilya Sutskever.
\newblock Zero-shot text-to-image generation.
\newblock In {\em International Conference on Machine Learning}, pages 8821--8831. PMLR, 2021.

\bibitem[\protect\citeauthoryear{Rumelhart \bgroup \em et al.\egroup }{1986}]{rumelhart1986learning}
David~E Rumelhart, Geoffrey~E Hinton, and Ronald~J Williams.
\newblock Learning representations by back-propagating errors.
\newblock {\em nature}, 323(6088):533--536, 1986.

\bibitem[\protect\citeauthoryear{Scellier and Bengio}{2017}]{scellier2017equilibrium}
Benjamin Scellier and Yoshua Bengio.
\newblock Equilibrium propagation: Bridging the gap between energy-based models and backpropagation.
\newblock {\em Frontiers in computational neuroscience}, 11:24, 2017.

\bibitem[\protect\citeauthoryear{Silver \bgroup \em et al.\egroup }{2017}]{silver2017mastering}
David Silver, Julian Schrittwieser, Karen Simonyan, Ioannis Antonoglou, Aja Huang, Arthur Guez, Thomas Hubert, Lucas Baker, Matthew Lai, Adrian Bolton, et~al.
\newblock Mastering the game of go without human knowledge.
\newblock {\em nature}, 550(7676):354--359, 2017.

\bibitem[\protect\citeauthoryear{Sporns and Bullmore}{2014}]{sporns2014connections}
Olaf Sporns and Edward~T Bullmore.
\newblock From connections to function: the mouse brain connectome atlas.
\newblock {\em Cell}, 157(4):773--775, 2014.

\bibitem[\protect\citeauthoryear{STEGER and WORMALD}{1999}]{STEGER_WORMALD_1999}
A.~STEGER and N.~C. WORMALD.
\newblock Generating random regular graphs quickly.
\newblock {\em Combinatorics, Probability and Computing}, 8(4):377–396, 1999.

\bibitem[\protect\citeauthoryear{Vaswani \bgroup \em et al.\egroup }{2017}]{vaswani2017attention}
Ashish Vaswani, Noam Shazeer, Niki Parmar, Jakob Uszkoreit, Llion Jones, Aidan~N Gomez, {\L}ukasz Kaiser, and Illia Polosukhin.
\newblock Attention is all you need.
\newblock {\em Advances in neural information processing systems}, 30, 2017.

\bibitem[\protect\citeauthoryear{Watts and Strogatz}{1998}]{watts1998collective}
Duncan~J Watts and Steven~H Strogatz.
\newblock Collective dynamics of ‘small-world’networks.
\newblock {\em nature}, 393(6684):440--442, 1998.

\bibitem[\protect\citeauthoryear{Waxman}{1988}]{12889}
B.M. Waxman.
\newblock Routing of multipoint connections.
\newblock {\em IEEE Journal on Selected Areas in Communications}, 6(9):1617--1622, 1988.

\bibitem[\protect\citeauthoryear{White \bgroup \em et al.\egroup }{1986}]{white1986structure}
John~G White, Eileen Southgate, J~Nichol Thomson, Sydney Brenner, et~al.
\newblock The structure of the nervous system of the nematode caenorhabditis elegans.
\newblock {\em Philos Trans R Soc Lond B Biol Sci}, 314(1165):1--340, 1986.

\bibitem[\protect\citeauthoryear{Winding \bgroup \em et al.\egroup }{2023}]{winding2023connectome}
Michael Winding, Benjamin~D Pedigo, Christopher~L Barnes, Heather~G Patsolic, Youngser Park, Tom Kazimiers, Akira Fushiki, Ingrid~V Andrade, Avinash Khandelwal, Javier Valdes-Aleman, et~al.
\newblock The connectome of an insect brain.
\newblock {\em Science}, 379(6636):eadd9330, 2023.

\end{thebibliography}

\end{document}